\pdfoutput=1

\documentclass[11pt]{article}

\usepackage[preprint]{acl}

\usepackage{dblfloatfix}

\usepackage{times}
\usepackage{latexsym}

\usepackage[T1]{fontenc}

\usepackage[utf8]{inputenc}

\usepackage{microtype}

\usepackage{inconsolata}

\usepackage{graphicx}

\usepackage{booktabs}
\usepackage{multirow}
\usepackage{amsmath}
\usepackage{bbm}
\usepackage[capitalise]{cleveref}
\usepackage{lipsum}
\usepackage{colortbl}
\usepackage{xcolor}
\usepackage{adjustbox}

\definecolor{gray1}{gray}{0.85}
\definecolor{gray2}{gray}{0.75}

\title{Cramming 1568 Tokens into a Single Vector and Back Again:\\ Exploring the Limits of Embedding Space Capacity}

\author{
Yuri Kuratov$^{1,2}$\quad
Mikhail Arkhipov$^3$\quad
Aydar Bulatov$^{1,2}$\quad
Mikhail Burtsev$^4$\\\\
$^1$AIRI, Moscow, Russia\\
$^2$Neural Networks and Deep Learning Lab, MIPT, Dolgoprudny, Russia \\
$^3$Independent Researcher, Amsterdam, Netherlands \\
$^4$London Institute for Mathematical Sciences, London, UK\\
\\
 \small{
   \textbf{Correspondence:} \href{mailto:yurii.kuratov@phystech.edu}{yurii.kuratov@phystech.edu}
 }
}

\begin{document}
\maketitle
\begin{abstract}
    A range of recent works addresses the problem of compression of sequence of tokens into a shorter sequence of real-valued vectors to be used as inputs instead of token embeddings or key-value cache. These approaches are focused on reduction of the amount of compute in existing language models rather than minimization of number of bits needed to store text. Despite relying on powerful models as encoders, the maximum attainable lossless compression ratio is typically not higher than x10. This fact is highly intriguing because, in theory, the maximum information capacity of large real-valued vectors is far beyond the presented rates even for 16-bit precision and a modest vector size. In this work, we explore the limits of compression by replacing the encoder with a per-sample optimization procedure. We show that vectors with compression ratios up to x1500 exist, which highlights two orders of magnitude gap between existing and practically attainable solutions. Furthermore, we empirically show that the compression limits are determined not by the length of the input but by the amount of uncertainty to be reduced, namely, the cross-entropy loss on this sequence without any conditioning.
The obtained limits highlight the substantial gap between the theoretical capacity of input embeddings and their practical utilization, suggesting significant room for optimization in model design.
\end{abstract}

\section{Introduction}

Most large language models (LLMs) are built on the Transformer architecture~\citep{vaswani2017attention} and have demonstrated remarkable performance as their parameters scale~\citep{radford2019gpt2,brown2020language,kaplan2020scaling,chinchilla_scaling}. As model sizes increase, so does the dimensionality of their input embeddings. However, despite this growth, each embedding still represents only a single token, e.g., for a series of Llama models embeddings size is growing from 2,048 in 1B-parameter models to 16,384 float numbers in 405B-parameter models~\citep{grattafiori2024llama}. Remarkably, even a 2,048-dimensional vector of 16-bit floats has a theoretical capacity of 32,768 bits, which is sufficient to encode roughly 1,931 tokens from a vocabulary of size 128,256. \emph{This observation motivates us to explore whether language models can utilize the latent capacity of input vectors more effectively, potentially encoding and processing multiple tokens with a single vector.}

\begin{figure}[t]
    \centering
    \includegraphics[width=0.8\columnwidth]{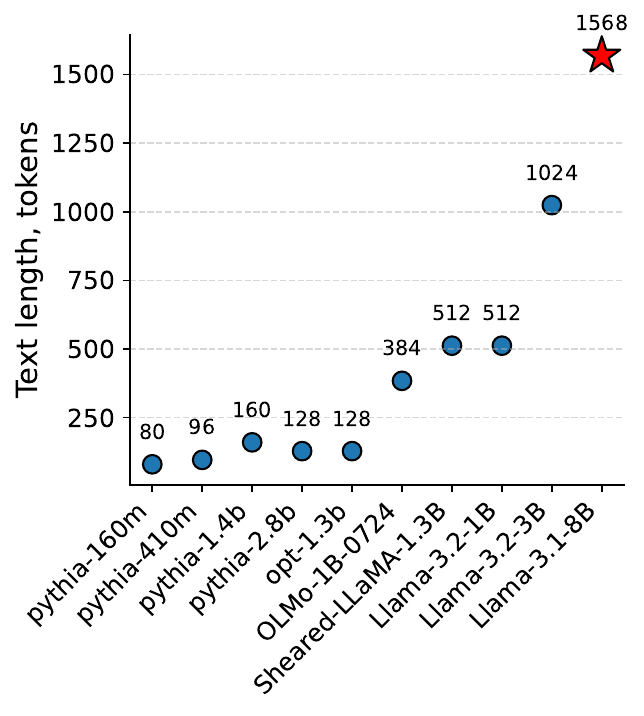}
    \caption{
    \textbf{How many tokens fit into a single input vector?} We estimate maximum number of tokens that can be decoded from a single input vector across various language models. 
    }
    \label{fig:results_brief}
\end{figure}

Encoding multiple tokens or even entire texts into a compact latent representation has been a longstanding challenge in natural language processing. It includes approaches, such as sentence embeddings~\citep{pmlr-v32-le14,NIPS2015_f442d33f,cer2018USE,wang2024e5multilingual} for semantic search and retrieval, and text autoencoders~\citep{bowman-etal-2016-generating,pmlr-v48-miao16,montero-etal-2021-sentence}, aimed to capture the essential meaning of texts in a dense representations.

In the context of LLMs, the challenge of encoding prompts and long contexts is particularly important because of the quadratic computational cost of the self-attention mechanism in Transformers. Several works have explored the possibility of replacing token-based prompts with a smaller set of dense vectors~\citep{lester2021power,li2021prefix,gao2024selfcp,li2024500xcompressor}, thereby shortening the input sequence and reducing the computational budget. These methods have demonstrated token-to-vector lossy compression ratios on the order of x500 with 8B-parameter models, indicating that it is possible to retain the critical information in a significantly reduced number of vectors. However, lossless compression is still limited by approximately factor of 10.

In memory-augmented architectures~\citep{weston2014memory,sukhbaatar2015endtoend, burtsev2021memory}, these embeddings act as additional storage or as a recurrent state for passing information between time steps~\citep{dai2019transformerxl,rmt_2022,chevalier2023adapting,behrouz2024titans}, essentially serving as an episodic memory. Moreover, recent approaches have explored the power of latent space reasoning~\citep{hao2024coconut} where high-capacity embeddings enable models to perform complex multi-step tasks directly in latent space. Consequently, the capacity of these vectors is crucial not only for efficient input representation, but also for increasing the overall expressiveness and computational power of models~\citep{merrill2023parallelism,strobl2024formal, sanford2024representational}.
Better understanding of the latent capacity of input vectors, could significantly help to improve encoding and retrieving of contextual information, episodic memory, as well as complex reasoning within large language models.

In this work, we investigate the limits of such input representations, exploring their capacity to encode and reconstruct long text sequences. By systematically quantifying how much additional information these vectors can capture, we provide insights into the efficiency and potential of latent representations in LLMs. Our main contributions:

1. We empirically study  capacity limits of LLM's input representations by compressing texts into trainable \texttt{[mem]} vectors.

2. We establish a direct connection between the latent capacity of input vectors and text cross-entropy, providing a quantitative measure of the information each vector can encode.

3. We show that the capacity limits remain consistent across different text lengths and domains, including natural text and random word sequences. 

4. We introduce a set of metrics that decouple the capacity of trainable input vectors from the language model's inherent prediction abilities. Using these metrics, we demonstrate a nearly linear scaling of compression capacity with the number of trainable vectors (e.g., Llama-3.2-1B compresses 7,168 tokens into just 16 vectors).

Our code is available at this \href{https://github.com/yurakuratov/hidden_capacity}{URL}.


\section{Related Work}




Approaches to compressing LLM context into a shorter sequence of input or KV-cache vectors are explored for various purposes, yet no standardized terminology or unified methodology has emerged. 
\paragraph{Context compression.} One application for input compression is connected with efficient processing of long contexts with LLMs. RMT~\cite{rmt_2022} and AutoCompressors~\cite{chevalier2023adapting} train the whole language model in a recurrent manner to compress the information from input segments to summary vectors and later reuse them to solve long-context tasks. ICAE~\citet{ge2023incontext} uses an autoencoder architecture with a frozen LLM as a decoder and adapt the same LLM for the encoder using LoRA~\cite{hu2022lora}. The resulting pipeline is pretrained using autoencoding and language modeling objectives, and then finetuned for language tasks, achieving the effective compression rate of x4. SelfCP~\cite{gao2024selfcp} uses the base LLM itself as a compressor using a trainable adapter to aggregate compressed states across multiple segments. 500xCompressor~\cite{li2024500xcompressor} extends the autoencoding approach with layer-wise connections and additional language pretraining tasks, exploring compression ratios up to x480, though at the cost of substantial quality degradation. 
Alternative approaches aim to compress KV activations instead of input tokens. Some methods achieve this by estimating token relevance, either through training-free~\cite{zhang2023h2o, li2024snapkv} or training-based approaches~\cite{qin2023nugget, qin2024dodo}, to prune irrelevant tokens and focus computation on the most informative ones. These strategies can yield high-quality but lossy compression with ratios up to x20. This result can be improved by finetuning models to leverage the resulting cache representations more effectively. This way, KV-Distill~\cite{chari2025kv} can reduce cache size up to 100 times with negligible loss in QA performance. 
In contrast, our method, applied to models of comparable size (up to 8 billion parameters), demonstrates that a compression rate of x1568 can be achieved without any loss in reconstruction quality.

\paragraph{Prompt compression.} Another line of work targets prompts compression to reduce inference costs. Gist tokens~\cite{mu2023learning} are prompt representations compressed by the LLM itself, finetuned with a special mask. Gisting allows to achieve prompt compression rate up to x26 with only minor loss in model performance. LLMLingua~\cite{jiang2023llmlingua} decouples the compression operation from the LLM and introduces a coarse-to-fine prompt compression strategy with a budget controller and token-level iterative compression, achieving up to x20 compression with negligible performance loss.~\citet{jiang2023longllmlingua2, pan2024llmlingua} extend this framework to long contexts, improving information retention through data distillation. 
Additionally,~\citet{morris-etal-2023-text,morris2023language} show that the original text can be reconstructed not only from its embeddings but even from the LM's predictions.
In the current work
we apply a per-sample optimization process instead to explore the fundamental limits of compression
and establish upper bounds on compression rates that far exceed prior work.

\paragraph{LLM-based lossless compression pipelines.} Language models have also been investigated for lossless text compression.\footnote{\citet{deletang2024language} explore data compression (texts to sequences of bits) using Arithmetic Coding and LLMs. This line of research is orthogonal to ours, as the sequence of bits cannot be passed as an input to an LLM. We analyze our approach from the data compression perspective in \cref{app:data_compression}.}
LLMZip~\cite{valmeekam2023llmzip} improves standard compression by ranking candidates using next-token probabilities, while FineZip~\cite{mittu2024finezip} accelerates compression through fine-tuning and dynamic context management for better efficiency.
The capabilities of LLMs in compression pipelines can be measured in bits-per-token over a representative textual corpus. \citet{huang2024compression, guo2024ranking} provide such measurements for public LLMs and establish the connection between compression rate and model performance, measured by diverse benchmark scores. 
Unlike these methods, we do not rely on external compression algorithms. Instead, we achieve lossless compression using only the LLM itself, providing both theoretical insights and practical demonstrations of compression limits.

\paragraph{Trainable tokens.} Some works use the trainable input tokens in other ways. \citet{burtsev2021memory} uses memory tokens as additional representation storage, \citet{beltagy2020longformer, zaheer2020big} use similar global tokens to enhance long-range information flow. \citet{li2021prefix,lester2021power,gao2024selfcp,liu2022ptuning} explore trainable soft prompts
as an alternative to finetuning model weights. 
Our findings about the representation capacity can represent the potential efficiency limits of such methods, based on how far the model behavior can be changed using trainable
tokens.

\section{Method}

\begin{figure}[ht]
    \centering
    \includegraphics[width=0.7\columnwidth]{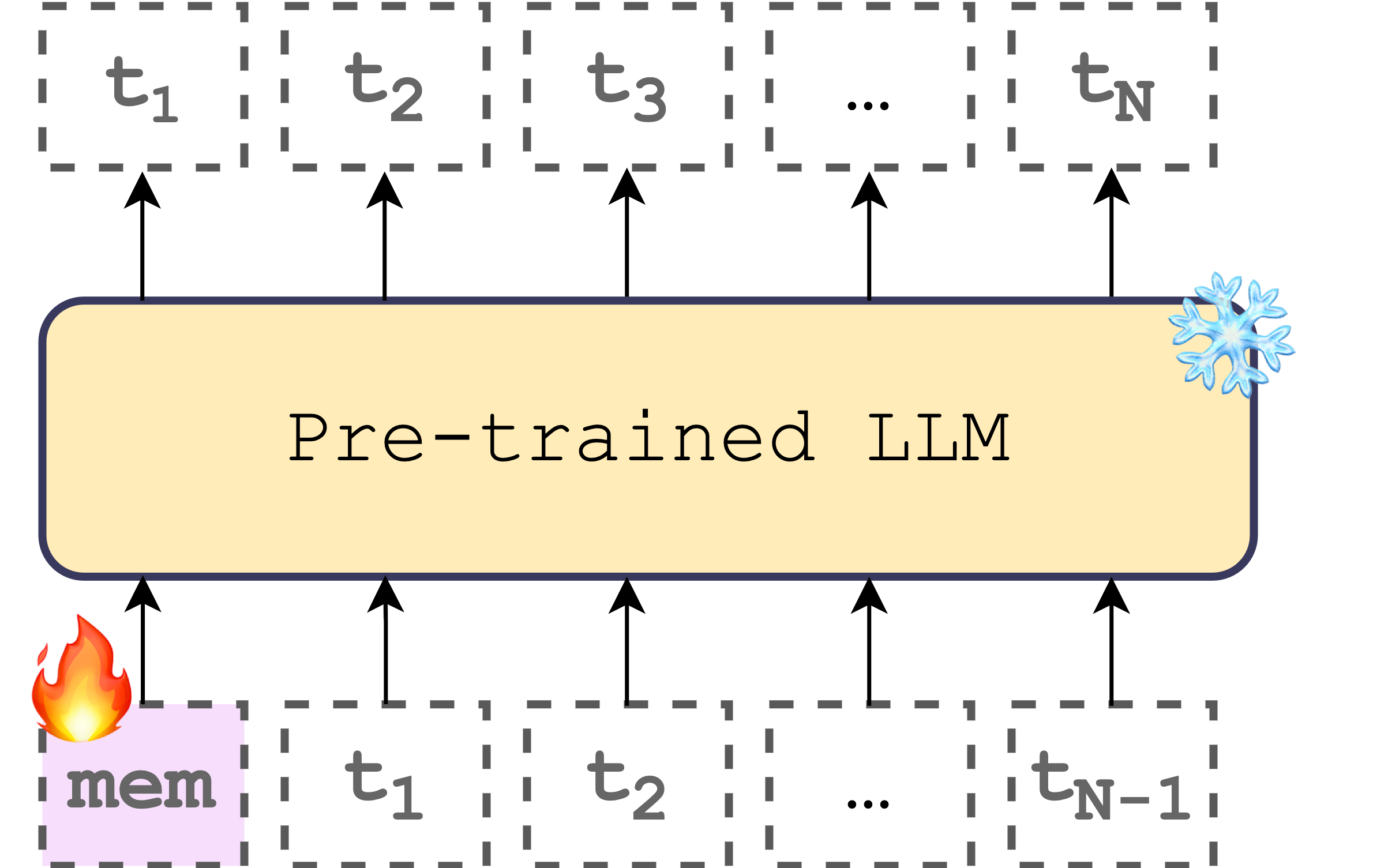}
    \caption{\textbf{Compressing text into a \texttt{[mem]} vector.} The pre-trained LLM is frozen, and we only finetune one or multiple \texttt{[mem]} vectors to decode the sequence of tokens $[t_1, t_2, \ldots, t_N]$. \texttt{[mem]} vectors are trained for each text separately.}
    \label{fig:compression_schema}
\end{figure}

We propose a simple approach for compressing a sequence of tokens into a small set of "memory" vectors. Then with this method we analyze how many tokens can be stored and decoded from a small set of resulting vectors. \cref{fig:compression_schema} provides an overview of our setup.

Trainable \texttt{[mem]} vectors are inspired by Memory Transformers~\cite{burtsev2021memory}, but here these vectors are designed to encode an entire text sequence. The training method is similar to Prompt Tuning~\cite{lester2021power}, with only a set of special input embeddings optimized while all parameters of the language model are frozen.

Formally, given a token sequence $[t_1,t_2,\dots, t_N]$, we introduce a set of trainable vectors $\texttt{[mem]} = [m_1,\dots,m_K]$ that are prepended to the text. 
These \texttt{[mem]} vectors are optimized to encode $[t_1,t_2,\dots, t_N]$.
During training, the frozen language model processes $[m_1,\dots,m_K,t_1,t_2,\dots, t_i]$ as the input context for predicting next token $t_{i+1}$. The \texttt{[mem]} vectors are optimized by minimizing the standard next-token prediction cross-entropy loss. As a result, each text sequence is associated with a unique set of \texttt{[mem]} vectors. At inference time, we start generation with the learned \texttt{[mem]} tokens and let LM to decode the text.

Let's estimate an upper bound on the number of tokens that can be generated from a single input vector by a language model. The input vector has a dimension $d_{model}$, with each element represented by $b$ bits, so that the total information content is approximately $d_{model} \times b$ bits. Given a vocabulary of size $|\mathcal{V}|$, where each token carries at most $\log_2 |\mathcal{V}|$ bits of information, the maximum number of tokens $L$ that can be generated is bounded by:
\begin{equation}
\label{eq:theoretical_capacity}
     L \leq \frac{d_{model} \times b}{\log_2 |\mathcal{V}|}.
\end{equation}
Our goal is to quantify the capacity 
of trainable input vectors (denoted as \texttt{[mem]}) in terms of the amount of information they can encode and later 
decode. From an information-theoretic standpoint, we interpret this capacity as the ability to reduce uncertainty in the generated text. To this end, we define the following metrics.

\textbf{Decoding Capacity (in Tokens)}:  
    From an information-theoretic perspective, this metric represents the maximum number of tokens that can be reliably reconstructed from the compressed representation in the \texttt{[mem]} vectors. It is defined as the longest text sequence length \(L\) for which the token-level accuracy exceeds a predefined threshold:
    \begin{multline}
    \label{eq:L_max}
    L_{\text{max}} = \max \big\{L \mid \\
    \text{Acc}\big(\text{LM}(t_{[1:L]} \mid \texttt{[mem]})\big) > \text{thr} \big\},
    \end{multline}
    here, \emph{Acc} is computed via teacher-forcing when decoding text both with and without the \texttt{[mem]} vectors. This measure reflects the effective storage limit (in tokens) imposed by the fixed capacity of the memory vector.

\textbf{Token Gain}:  
    This metric estimates the additional number of tokens that can be correctly decoded due to the presence of the \texttt{[mem]} vector, relative to the baseline performance of the language model (LM) without it. Formally, if \(C^{\text{LM+[mem]}}_{\text{tokens}}\) is the count of tokens correctly predicted when using the memory vector and \(C^{\text{LM}}_{\text{tokens}}\) is the count without it, then the gain is given by
    \begin{align}
    \label{eq:capacity_tokens}
    C_{\text{tokens}} &= C^{\text{LM+[mem]}}_{\text{tokens}} - C^{\text{LM}}_{\text{tokens}} \notag \\
    &= \sum_{i=1}^{N} \mathbbm{1}\Big( t_i = \text{LM}(t_{[1:i-1]} \mid \texttt{[mem]}) \Big) \notag \\
    &\quad - \sum_{i=1}^{N} \mathbbm{1}\Big( t_i = \text{LM}(t_{[1:i-1]}) \Big).
    \end{align}
    Viewed through an information-theoretic lens, this difference quantifies the number of tokens' worth of information (i.e., discrete units) that the memory vector adds to the decoding process.

  \textbf{Information Gain}:  
    Cross-entropy measures the uncertainty or the average number of bits required to encode a sequence under a given model. The \emph{Information Gain} quantifies how much the \texttt{[mem]} vector reduces this uncertainty. Let $H_{\text{LM}} = H\big( P_\theta(t_{[1:N]}) \big)$
    be the cross-entropy (in bits) when decoding without the memory vector, and $
    H_{\text{LM+[mem]}} = H\big( P_\theta(t_{[1:N]} \mid \texttt{[mem]}) \big)$
    be the cross-entropy with the memory vector.
    Then, the reduction is given by
    \begin{align}
    \label{eq:capacity_entropy}
        \text{CE-reduction} &= C_{\text{H}} = H_{\text{LM}} - H_{\text{LM+[mem]}}.
    \end{align}
    This measures how many fewer bits are needed to represent the text, thus reflecting the additional information provided by the memory vector.

Collectively, these metrics enable us to characterize the capacity of the trainable input vectors both in terms of discrete tokens (\(C_{\text{tokens}}\)) and entropy (\(C_{\text{H}}\)), while \(L_{\text{max}}\) provides an upper bound on the length of text that can be accurately reconstructed. In our experiments, these measures are computed over a curated set of texts and averaged to obtain robust estimates. We note that the absolute values of \emph{Information Gain} depend on the underlying vocabulary, and therefore should not be directly compared across models with different vocabularies.

\section{Experiments and Results}

We evaluate capacity of trainable input vectors of the same size as dimension of input embeddings for different language models on texts from different sources.

\paragraph{Models}
We use models from Pythia suite (160M, 410M, 1.4B, 2.8B)~\cite{biderman2023pythia}, OPT-1.3B~\cite{zhang2022opt}, OLMo-1B~\cite{groeneveld-etal-2024-olmo}, 
Sheared-LLaMA-1.3B~\cite{xia2024sheared}, Llama-3 models (1B, 3B, 8B)~\cite{grattafiori2024llama}, and Mamba (130M, 370M, 790M, 1.4B)~\cite{gumamba}. List of all used models with links to HuggingFace Hub are in Appendix~\ref{app:models}.

\paragraph{Data}

\begin{table*}[t]
\centering
\renewcommand{\arraystretch}{1.2}
\resizebox{\linewidth}{!}{%
\begin{tabular}{llllllll}
\toprule
       & & \textbf{Pythia-160M} & \textbf{Pythia-410M} & \textbf{Pythia-1.4B} & \textbf{Llama-3.2-1B} & \textbf{Llama-3.2-3B} & \textbf{Llama-3.1-8B} \\
\midrule
\multirow{3}{*}{\textbf{PG-19}} & Max, tokens &                      80 &                      96 &                       160 &                       512 &                      1024 &                      1568 \\
      &  \cellcolor{gray1}Gain, tokens & \cellcolor{gray1}70.9\tiny$\pm$11.0 & \cellcolor{gray1}81.3\tiny$\pm$12.0 & \cellcolor{gray1}158.0\tiny$\pm$29.1 & \cellcolor{gray1}426.2\tiny$\pm$79.2 & \cellcolor{gray1}720.3\tiny$\pm$80.2 & \cellcolor{gray1}1094.1\tiny$\pm$127.6 \\
       & \cellcolor{gray2}Information Gain & \cellcolor{gray2}396.4\tiny$\pm$46.0 & \cellcolor{gray2}431.4\tiny$\pm$51.6 & \cellcolor{gray2}792.8\tiny$\pm$143.4 & \cellcolor{gray2}2119.9\tiny$\pm$364.8 & \cellcolor{gray2}3292.2\tiny$\pm$320.0 & \cellcolor{gray2}4865.7\tiny$\pm$546.6 \\
\cline{1-8}
\multirow{3}{*}{\textbf{Fanfics}} & Max, tokens &                      80 &                      96 &                       192 &                       512 &                      1024 &                      1568 \\
       & \cellcolor{gray1}Gain, tokens & \cellcolor{gray1}70.9\tiny$\pm$10.5 & \cellcolor{gray1}81.2\tiny$\pm$11.6 & \cellcolor{gray1}152.9\tiny$\pm$28.0 & \cellcolor{gray1}449.6\tiny$\pm$83.7 & \cellcolor{gray1}734.1\tiny$\pm$85.0 & \cellcolor{gray1}1071.8\tiny$\pm$168.6 \\
       & \cellcolor{gray2}Information Gain & \cellcolor{gray2}378.1\tiny$\pm$45.9 & \cellcolor{gray2}429.8\tiny$\pm$46.2 & \cellcolor{gray2}776.9\tiny$\pm$132.5 & \cellcolor{gray2}2213.8\tiny$\pm$365.8 & \cellcolor{gray2}3354.5\tiny$\pm$344.9 & \cellcolor{gray2}4768.9\tiny$\pm$622.6 \\
\cline{1-8}
\multirow{3}{*}{\textbf{Random}} & Max, tokens &                      65 &                      72 &                       139 &                       316 &                       460 &                       792 \\
       & \cellcolor{gray1}Gain, tokens & \cellcolor{gray1}61.3\tiny$\pm$6.6 & \cellcolor{gray1}76.9\tiny$\pm$8.7 & \cellcolor{gray1}144.4\tiny$\pm$17.5 & \cellcolor{gray1}294.9\tiny$\pm$64.8 & \cellcolor{gray1}456.9\tiny$\pm$72.1 & \cellcolor{gray1}623.2\tiny$\pm$97.3 \\
       & \cellcolor{gray2}Information Gain & \cellcolor{gray2}500.8\tiny$\pm$38.9 & \cellcolor{gray2}630.4\tiny$\pm$65.2 & \cellcolor{gray2}1108.2\tiny$\pm$136.2 & \cellcolor{gray2}2265.2\tiny$\pm$498.7 & \cellcolor{gray2}3382.6\tiny$\pm$585.2 & \cellcolor{gray2}4541.2\tiny$\pm$758.6 \\
\bottomrule
\end{tabular}
}%
\caption{
  \textbf{Compression capacity across different text sources and models.}
  We report \emph{Decoding Capacity (in Tokens)} ("Max, tokens" in the Table), \emph{Token Gain}, and \emph{Information Gain} for texts from \emph{PG-19}, \emph{fanfics}, \emph{random}.
  Notably, \emph{Information Gain} remains similar across all text sources for each model (except \emph{random} for Pythia).
  For \emph{PG-19} and \emph{fanfics}, LMs leverage their ability to predict natural language, so the \emph{Decoding Capacity (in Tokens)} generally exceeds the \emph{Token Gain}. Furthermore, we find no evidence that the models benefit from potentially having PG-19 in their pre-training data, as their performance on \emph{PG-19} is not significantly better than on \emph{fanfics} published after October 2024.
  In contrast, random text offers no predictable structure, making these two metrics nearly identical. This allows us to distinguish how many tokens model can predict by itself compared to decoding from trainable input vector.
  Larger models consistently show greater compression capacity across all metrics.
}
\label{tab:compression_pg_fanfics_random}
\end{table*}
As a source of texts for compression, we use texts from the PG-19 dataset~\cite{rae2019compressive}, which consists of books extracted from the Project Gutenberg library. Given that PG-19 is publicly available and contains books, it is highly plausible that these texts were included in the pre-training data of LLMs. Notably, PG-19 is part of the Pile dataset~\cite{pile}, which was used to train Pythia models.

To assess the compression of texts that models have not encountered during pre-training, we collected fanfiction texts published online after October 2024 from the AO3 fanfics library\footnote{\url{https://archiveofourown.org/}}. Details of this collection process are provided in Appendix~\ref{app:data_fanfics}.

Both the \emph{PG-19} and \emph{fanfics} consist of natural language texts, where language models can predict some tokens based on prior context and model parameters. To isolate the capacity of the trainable input vectors without the influence of the knowledge of natural language by language model itself, we also employed \emph{random} texts. \emph{Random} texts were generated by randomly sampling words from the top 100,000 words from the GloVe vocabulary\footnote{\url{https://nlp.stanford.edu/data/glove.6B.zip}}.

We train only a set of $M$ vectors that are prepended to the model's input. In most of the experiments, we use only one trainable vector, if not stated otherwise.

\subsection{Decoding Capacity of a Single Vector}
\label{sec:compression_in_tokens}

We find that a \emph{single} trainable vector can enable language models to produce surprisingly long, \emph{targeted} text sequences. We estimate \emph{Decoding Capacity (in Tokens)} (\cref{eq:L_max}) on 50 texts from PG-19 for each length.
We set a token-level accuracy threshold of $0.99$ and evaluate across the following length grid: [64, 80, 96, 128, 160, 192, 256, 384, 512, 768, 1024, 1280, 1568, 2048, 2560, 3072].

\cref{fig:results_brief} presents the results for the evaluated models. Notably, Llama-3.1-8B can accurately reconstruct texts of up to $1568$ tokens from just a \emph{single} input vector. Interestingly, among models with around 1B parameters (Pythia-1.4B, OPT-1.3B, OLMo-1B, Sheared-LLaMA-1.3B, and Llama-3.2-1B) we observe compressive capacity that ranges from $128$ to $512$ tokens. Pythia-2.8b, despite its larger size, has poor compression of just $128$ tokens compared to smaller 1B models.


\subsection{Memorization, Natural Language Understanding and Episodic Memory}

Generation from the \texttt{[mem]} vector involves combining information from both the pre-trained language model parameters and memory about text specific sequence. To analyze contributions of these different types of memory, we use  \emph{Token Gain} (\cref{eq:capacity_tokens}) which measure the extra number of tokens predicted correctly, and \emph{Information Gain} (\cref{eq:capacity_entropy}) showing the decrease in cross-entropy when decoding from memory vector. In contrast to \emph{Decoding Capacity}, these two metrics more directly isolate the capacity contributed by the \texttt{[mem]} vector itself.

\begin{figure*}[t]
  \centering
  \includegraphics[width=0.95\linewidth]{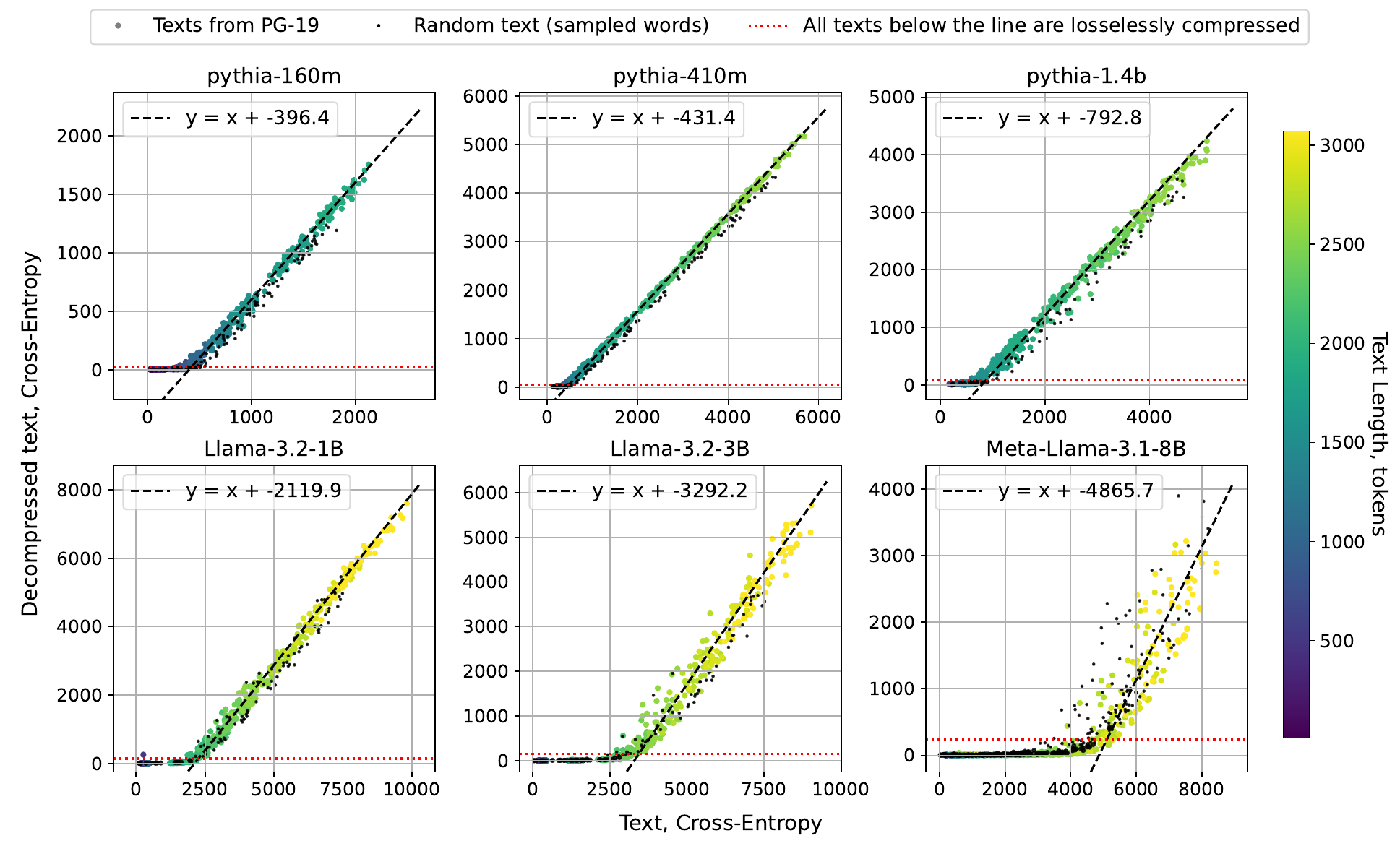} 
  \caption{
  \textbf{Information gain of text compression to \texttt{[mem]} vector doesn't depend on language understanding capabilities of models.} Compression results for various language models show the relationship between the cross-entropy (CE) of the original and decompressed texts. If the text CE falls below a model-specific threshold (red line), the text is losslessly compressed. This value is a input vector capacity in terms of entropy (\emph{Information Gain}, $C_H$).
  For texts that are not perfectly compressed, the compression process reduces their CE to a consistent, model-specific value (bias of the black dashed line).
Larger models (e.g., Llama-3.1-8B) can handle longer texts before reaching the compression threshold, due to their greater capacity compared to smaller models (e.g., Pythia-160M). This behavior holds for both natural texts (\emph{PG-19}) and unnatural \emph{random} texts consisting of random word sequences.
  }
  \label{fig:entropy}
\end{figure*}

In addition to texts from \emph{PG-19} that may have been seen by LMs during pre-training, we consider: (1) texts from ~\emph{fanfics} to factor out memorization as they were published after release of the models, and (2)~\emph{random} sequenses of words to exclude learned natural language understanding capabilities.

\emph{Decoding Capacity (in Tokens)} for texts from \emph{PG-19} and \emph{fanfics} was evaluated on the following length grid: [64, 80, 96, 128, 160, 192, 256, 384, 512, 768, 1024, 1280, 1568, 2048, 2560, 3072].

\cref{tab:compression_pg_fanfics_random} summarizes the results for each model and text source. We have two main observations. 
The metrics for both \emph{PG-19} and \emph{fanfics} are remarkably similar across all models tested. This similarity implies that the presence of \emph{PG-19} in the pre-training data does not provide much of an advantage. Thus, compression performance does not appear to be driven by direct memorization of the dataset. Notably, even for \emph{random} texts, larger models such as Llama-3.1-8B still exhibit substantial compressing power, reliably reconstructing sequences of up to 792 tokens. This result demonstrates the impressive capacity of learnable input embeddings to control LLM generation. In particular, a single learned vector is sufficient to guide generation of nearly 800 random tokens.

A key takeaway from these results is that the model's compression ability does not depend on familiarity with specific texts or knowledge of natural language gained during pre-training. Instead, the single trainable vector itself provides language agnostic substantial capacity, allowing to store completely novel texts or random sequences of words.

\begin{figure*}[t]
  \centering
  \includegraphics[width=0.95\linewidth]{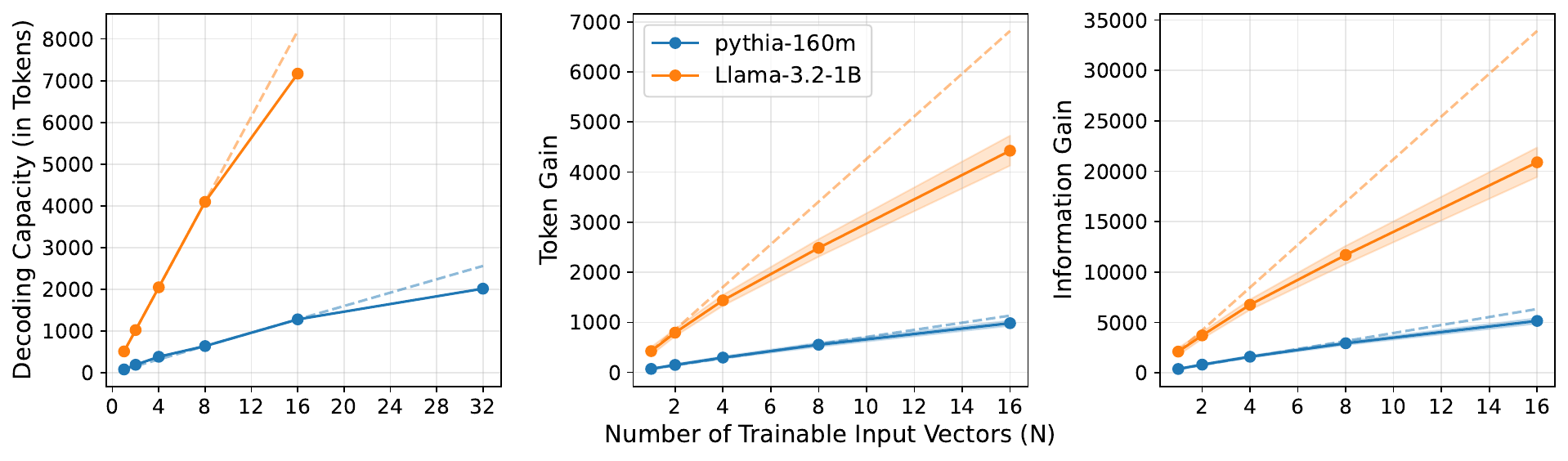}
  \caption{
  \textbf{Compression scales linearly with the number of trainable \texttt{[mem]} vectors.} 
  Dashed lines represent ideal linear scaling, and shaded regions indicate $\pm1$ std.
  Pythia-160m reaches its maximum input context length of 2048 tokens and can successfully encode texts of up to 2016 tokens into 32 \texttt{[mem]} input vectors. Llama-3.2-1B can perfectly decode texts of 7168 tokens from just 16 input vectors.
}
\label{fig:scale_mem}
\end{figure*}

\subsection{Sensitivity of Compression to Text Complexity}
\label{sec:compression_in_entropy}

Decoding capacity might depend on the complexity of the input text for a language model. In this section, we study how compression changes uncertainty of the model about the text.

For 50 text samples from the \emph{PG-19} at each target length (ranging from 8 up to 1568 tokens, and to 3072 for larger models) we measured cross-entropy both before ($H_{\text{LM}}$) and after ($H_{\text{LM+[mem]}}$) compression (see  \cref{eq:capacity_entropy}). 
Figure~\ref{fig:entropy} compares results across Pythia and LLama models, and full results for all models are provided in \cref{app:entropy_all_models}. 

In~\cref{fig:entropy}, the models demonstrate linear relationship between cross-entropy before and after compression for not perfectly compressible texts (i.e., lying above the red dotted line), indicating constant value of information gain (or, reduction in cross-entropy). Texts with cross-entropy lower than a model's information gain are perfectly reconstructed.

To verify that this also holds for arbitrary texts, we used \emph{random} word sequences and observed a similar pattern: as long as cross-entropy of a sample remains below the model-specific cutoff, it can be perfectly reconstructed. Notably, these random texts (black dots in~\cref{fig:entropy}) lie very close to the same linear trend as the \emph{PG-19} texts, showing that similar compression laws apply regardless of the nature of the sequence. Thus, $\texttt{[mem]}$ works as an episodic memory storing sequence specific information independent of natural language knowledge the model has.

\subsection{Scaling Compression with More Trainable Vectors}

To explore how compression scales with the number of input vectors $\texttt{[mem]} = [m_1,\dots,m_K]$ we use the same training process as before but for different numbers of trainable vectors, from 1 to 16 for the Llama-3.2-1B model and from 1 to 32 for Pythia-160M.

The results of this series of experiments are presented in~\cref{fig:scale_mem}, demonstrating that input vector capacity scales almost linearly with the number of trainable \texttt{[mem]} vectors. This trend holds consistently across all measures of capacity, whether expressed in terms of tokens or text entropy. In particular, Pythia-160M successfully decodes texts up to 2016 tokens in length using 32 \texttt{[mem]} vectors, effectively reaching its maximum context length. Similarly, LlaMA-3.2-1B achieves perfect reconstruction for sequences as long as 7168 tokens with just 16 input vectors. However, scaling behavior for LlaMA-3.2-1B deviates from the linear trend, suggesting potential inefficiencies in the compression process or inherent model limitations in exploiting an increasing number of input vectors for information storage and extraction.

Extrapolating from these trends, we estimate that an entire text such as "The Hobbit, or There and Back Again" (approximately 120,000 tokens) could be compressed into only 128 input vectors using Llama-3.1-8B and into 256 vectors using Llama-3.2-1B.

These results demonstrate that increasing the number of trainable \texttt{[mem]} vectors significantly enhances compression capacity, with linear scaling observed across the evaluated models. Notably, using a small number of additional vectors introduces minimal computational overhead while enabling the reconstruction of substantially longer texts.

\subsection{Embedding Capacity Utilization}
\label{sec:capacity_utilization}

\begin{figure}[t]
    \centering
    \includegraphics[width=1.0\columnwidth]{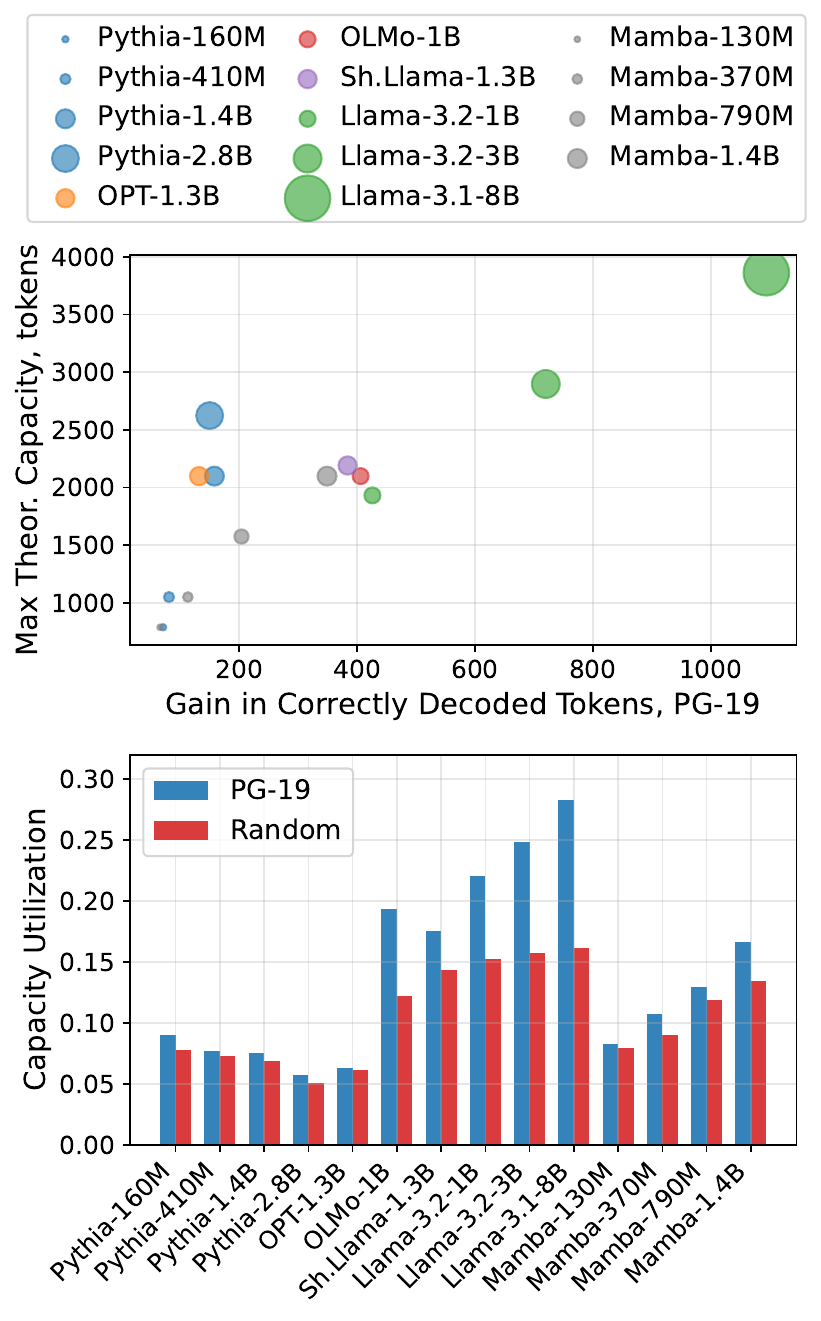}
    \caption{
    \textbf{Only fraction of learned input embedding information capacity can be utilized.} \textit{Top.} Maximum token capacity (see~\cref{eq:theoretical_capacity}) against gain in correctly decoded tokens shows differences in utilization of learned memory embedding for studied models.
    \textit{Bottom.} Capacity utilization for natural and random texts.
    } 
    \label{fig:capacity_utilization}
\end{figure}

To measure how effectively each model uses its input embedding space, we compare the empirically measured capacity in tokens (\emph{Token Gain}) to a theoretical maximum derived from embedding size and vocabulary size (see~\cref{eq:theoretical_capacity}). We define \emph{capacity utilization} as the ratio of these two quantities.

In~\cref{fig:capacity_utilization}~(top), when comparing all models with roughly 1B parameters, there are two groups: (1) older models (e.g., OPT and Pythia) show lower capacity utilization, whereas (2) newer models (e.g., Llama, ShearedLlama, Mamba, and OLMo) demonstrate higher utilization despite having the same theoretical capacity. This disparity indicates that the quality of pre-training (data, compute budget, improvements in architecture) influences the extent to which a model can exploit its input vectors 
capacity.

In~\cref{fig:capacity_utilization}~(bottom) we can see three groups: (1) and (2) same as on top plot with older and newer models, and (3) group of Mamba models. The Pythia models show an interesting trend: as model size increases, capacity utilization decreases. This pattern suggests that the larger Pythia models may be under-trained relative to their theoretical potential. In contrast, Llama and OLMo models show higher capacity utilization. Based on these observations, we hypothesize that capacity utilization could serve as an indicator of the pre-training status and guide further training.

For models within the Llama family, we observe that empirical capacity utilization rises steadily with model size (1B, 3B, 8B), most noticeable on natural texts from \emph{PG-19}.
This might be fully attributed to better language understanding, gained by the larger models during pre-training, but even on \emph{random} texts we observe a modest upward trend.
This result suggests that the overall number of parameters plays an important role in determining effective capacity not only via LM capabilities but also due to better utilization of embedding space for episodic information storage.

\subsection{Non-Transformer Models: Mamba}
Notably, the findings from the previous sections are not limited to Transformer-based language models. State-space Mamba models can also encode an entire text into a single \texttt{[mem]} vector and perfectly reconstruct the text when its cross-entropy falls below the model's capacity threshold, e.g., Mamba-1.4B can encode texts of $512$ tokens into a single \texttt{[mem]} vector, similar to other 1B models (Llama-3.2-1B, Sheared-LLaMa-1.3B). This demonstrates that obtained results are architecture-agnostic.

On capacity utilization (\cref{fig:capacity_utilization}), the Mamba models exhibit a consistent pattern across both panels that depends on model size. In the top plot, the increase in correctly decoded tokens is nearly linear with respect to each model's theoretical capacity. This implies that each additional parameter in the embedding space translates into greater decoding accuracy at a nearly constant rate. In the bottom plot, capacity utilization rises monotonically from the 130M to the 1.4B model. This is the opposite of the downward trend seen in Pythia, but aligns with Llama, and shows that larger Mambas progressively make better use of their learned input space. We give more details on Mamba results in~\cref{app:mamba,fig:entropy_all_models,tab:compression_pg_fanfics_random_all_models}).

\section{Discussion and Conclusions}

In this work, we introduced a simple yet effective way to compress entire text sequences into a small set of trainable \texttt{[mem]} vectors without any information loss. We used this method to analyze how far we can push the latent capacity of large language models compared to its theoretical limits.

By systematically evaluating different models, we find that a surprising amount of text can be compressed to a single token, and this capacity scales linearly with the number of tokens. 
This highlights significant potential in practical compression pipelines and long-context processing. We demonstrate that our compression outperforms neural models as a compression method, suggesting a more efficient approach to representing information. However, significantly more compute is needed due to optimization nature of the proposed method.

We establish a direct link between representation capacity and cross-entropy, showing that it remains independent of text length, domain, or familiarity. However, the exact model characteristics that determine capacity remain an open question. The hidden state dimension and model size play an important role along with general performance, however further analysis is required to determine the exact scaling laws for capacity.

Compression ability serves as a strong indicator of an LLM’s potential. Since transformers operate entirely within their representation space, its capacity fundamentally constrains reasoning, intermediate computations, and large-scale information processing. All textual and soft prompts ultimately reside in this space, meaning its limits define how effectively models can be steered and conditioned. By mapping these boundaries, we gain deeper insight into the fundamental constraints of current architectures and the possibilities for more powerful future models.

Moreover, our findings hold significant promise for memory-augmented architectures. The ability to compress long sequences into a compact set of memory vectors shows the way for integrating efficient external memory modules that can store and retrieve detailed episodic information, potentially enhancing reasoning, long-term dependency handling, and overall model performance. We believe that incorporating such optimized memory representations could lead to novel architectures that are both computationally efficient and more capable of complex information processing.

We believe our findings present an important stepping stone to understanding the limits of modern LLMs and building more powerful models in the future.

\section*{Limitations}

While our experiments push the boundaries of compression with LLMs and offer insights into their upper capacity limits, the nature of the obtained representations remains largely unclear.
We have analyzed the structure of the space of trained \texttt{[mem]} vectors in~\cref{app:compressed_vectors_analysis}, but more in-depth analysis is needed to determine the semantic properties of the vectors and their potential value in downstream tasks. 
Our findings are limited to Transformer-based and Mamba models with up to 8 billion parameters due to computational constraints. Investigating the representation space of larger models, as well as exploring alternative architectures such as recurrent and memory-augmented models, remains an important avenue for future research.
In our study with different text sources, we generate random text by sampling words from a dictionary. While this approach simplifies the analysis, it may slightly overestimate model capacity compared to sampling directly from a tokenizer’s vocabulary, as dictionary words can be split on multiple tokens by model. 

\section*{Broader Impact}
We train a set of \texttt{[mem]} vectors so that arbitrary texts can be accurately reconstructed from them. This process not only allows us to analyze the capacity of these vectors, but also demonstrates that any kind of text can be compressed into compact latent representations and later decoded. Such a capability may have far-reaching implications: it could lead to more efficient methods for storing and transmitting text, while also raising important considerations regarding the potential misuse of compressed information and issues related to data security, intellectual property, and altering the behavior of aligned models.

\section*{Acknowledgments}
We are thankful to SberDevices for granting us access to additional computational resources. This work was partially supported by the Ministry of Economic Development of the Russian Federation (code 25-139-66879-1-0003).

\bibliography{custom}

\clearpage

\appendix

\section{Models and Training Details}
\label{app:models}
We provide list of all models that we used in our experiments in~\cref{tab:hf_models}.

Trainable vectors are initialized randomly. We use the AdamW optimizer~\cite{loshchilov2018adamw} with a learning rate of $0.01$, $\beta_1$, and $\beta_2$ both set to $0.9$, and a weight decay of $0.01$. Training proceeds for a maximum of 5,000 steps, with early stopping if the text is compressed losslessly, i.e., achieving a token-level accuracy of 1.0. All models are loaded from the HuggingFace Transformers library with the PyTorch framework.

Each compression experiment was run on a single A100 80GB GPU. The time required to compress text using 5,000 optimization steps ranged from a dozen of seconds for small models and short contexts to 10--20 minutes for larger models and longer contexts. We used up to 4 GPUs to run several experiments in parallel.

\begin{table*}[h]
\centering
\resizebox{.9\linewidth}{!}{%
\begin{tabular}{llccc}
\toprule
\textbf{Model Name} & \textbf{Link to HuggingFace} & \textbf{Params (B)} & \textbf{Input Hidden Size} & \textbf{Vocabulary Size} \\
\midrule
Pythia-160M & \href{https://huggingface.co/EleutherAI/pythia-160m}{EleutherAI/pythia-160m} & 0.16 & 768 & 50304 \\
Pythia-410M & \href{https://huggingface.co/EleutherAI/pythia-410m}{EleutherAI/pythia-410m} & 0.41 & 1024 & 50304 \\
Pythia-1.4B & \href{https://huggingface.co/EleutherAI/pythia-1.4b}{EleutherAI/pythia-1.4b} & 1.4 & 2048 & 50304 \\
Pythia-2.8B & \href{https://huggingface.co/EleutherAI/pythia-2.8b}{EleutherAI/pythia-2.8b} & 2.8 & 2560 & 50304 \\
OPT-1.3B & \href{https://huggingface.co/facebook/opt-1.3b}{facebook/opt-1.3b} & 1.3 & 2048 & 50272 \\
OLMo-1B & \href{https://huggingface.co/allenai/OLMo-1B-0724-hf}{allenai/OLMo-1B-0724-hf} & 1.0 & 2048 & 50304 \\
Sheared-LLaMA-1.3B & \href{https://huggingface.co/princeton-nlp/Sheared-LLaMA-1.3B}{princeton-nlp/Sheared-LLaMA-1.3B} & 1.3 & 2048 & 32000 \\
Llama-3.2-1B & \href{https://huggingface.co/meta-llama/Llama-3.2-1B}{meta-llama/Llama-3.2-1B} & 1.0 & 2048 & 128256 \\
Llama-3.2-3B & \href{https://huggingface.co/meta-llama/Llama-3.2-3B}{meta-llama/Llama-3.2-3B} & 3.0 & 3072 & 128256 \\
Llama-3.1-8B & \href{https://huggingface.co/meta-llama/Llama-3.1-8B}{meta-llama/Llama-3.1-8B} & 8.0 & 4096 & 128256 \\
Mamba-130M & \href{https://huggingface.co/state-spaces/mamba-130m-hf}{state-spaces/mamba-130m-hf} & 0.13 & 768 & 50280 \\
Mamba-370M & \href{https://huggingface.co/state-spaces/mamba-370m-hf}{state-spaces/mamba-370m-hf} & 0.38 & 1024 & 50280 \\
Mamba-790M & \href{https://huggingface.co/state-spaces/mamba-790m-hf}{state-spaces/mamba-790m-hf} & 0.79 & 1536 & 50280 \\
Mamba-1.4B & \href{https://huggingface.co/state-spaces/mamba-1.4b-hf}{state-spaces/mamba-1.4b-hf} & 1.4 & 2048 & 50280 \\
\bottomrule
\end{tabular}
}
\caption{List of used language models and their parameters.}
\label{tab:hf_models}
\end{table*}

\section{Collecting Texts from the Fanfics Library}
\label{app:data_fanfics}

We used the AO3 fanfiction library \url{https://archiveofourown.org/} as a source of texts that were not present in the language models' pre-training data. To ensure novelty, we manually downloaded 21 fanfics from various fandoms (including Harry Potter, Star Wars, Transformers, Lord of the Rings, and others) that contained more than 20,000 words and were published after October 2024. 

We preprocessed the HTML pages to extract only the main text content, removing any irrelevant elements. We then sampled passages from these texts to evaluate the capacity of trainable input vectors. Throughout our experiments, we refer to this dataset as \emph{fanfics}.

From each of the \emph{PG-19} and \emph{fanfics}, we sampled texts and set their lengths to match the desired token counts. We ensured that each text began with complete sentences to maintain coherence. 
As a result, to estimate the capacity of the input vectors, we used 50 texts for each length.

\section{Non-Transformer Language Models: Mamba}
\label{app:mamba}
Mamba~\citep{gumamba} is a state space language model (SSM)~\citep{gu2022efficiently}, in contrast to attention-based Transformers. We applied the same compression procedure used for the Transformer models and evaluated the official Mamba checkpoints from the Hugging Face Hub (\cref{tab:hf_models}). We found that Mamba can also encode texts into a single \texttt{[mem]} vector and successfully reconstruct them whenever the text's cross-entropy falls below model's capacity threshold, indicating the effect is not unique to only Transformer models (\cref{fig:entropy_all_models,tab:compression_pg_fanfics_random_all_models}).

\section{Results of Evaluating Text Compression for All Models}
\label{app:entropy_all_models}
\begin{figure*}[t]
  \centering
  \includegraphics[width=\linewidth]{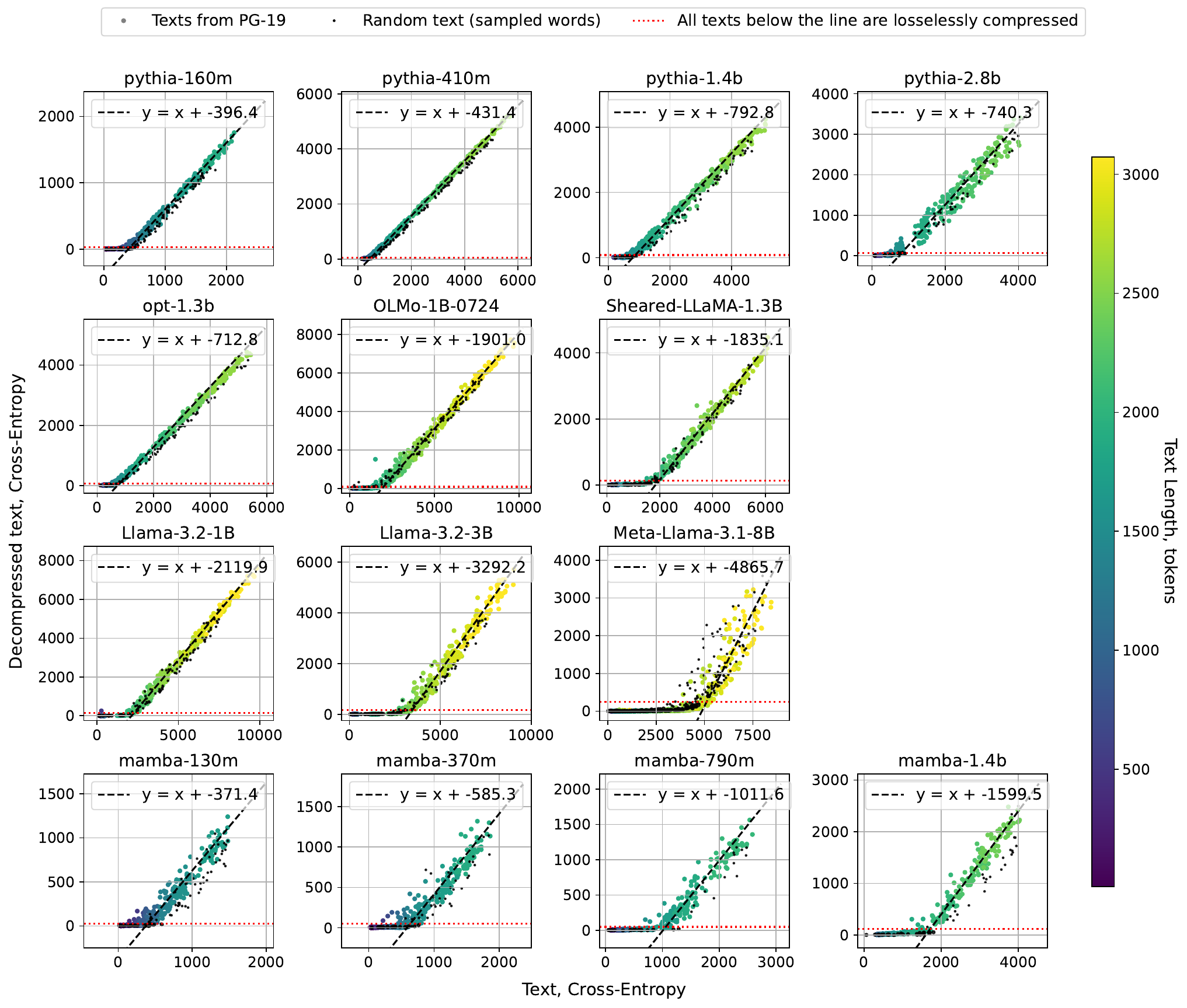} 
  \caption{
\textbf{Information gain of text compression to \texttt{[mem]} vector doesn't depend on language understanding capabilities of models.} Compression results for various language models show the relationship between the cross-entropy (CE) of the original and decompressed texts. If the text CE falls below a model-specific threshold (red line), the text is losslessly compressed. This value is a input vector capacity in terms of entropy (\emph{Information Gain}, $C_H$). 
  For texts that are not perfectly compressed, the compression process reduces their CE to a consistent, model-specific value (bias of the black dashed line).
Larger models (e.g., Llama-3.1-8B) can handle longer texts before reaching the compression threshold, due to their greater capacity compared to smaller models (e.g., Pythia-160M). This behavior holds for both natural texts (\emph{PG-19}) and unnatural \emph{random} texts consisting of random word sequences. The result is not limited to transformer-based architectures: non-transformer models, such as \emph{Mamba}, exhibit the same pattern.}
  \label{fig:entropy_all_models}
\end{figure*}

Here we provide results for all evaluated models in~\cref{fig:entropy_all_models,tab:compression_pg_fanfics_random_all_models}. Results are discussed in~\cref{sec:compression_in_entropy}.

\begin{table*}[t]
\centering
\renewcommand{\arraystretch}{1.2}
\resizebox{\linewidth}{!}{%
\setlength{\tabcolsep}{3pt}
\begin{tabular}{l |rrr| rrr| rrr}
\toprule
\multirow{2}{*}{\textbf{Model}} &
\multicolumn{3}{c}{\textbf{PG-19}} &
\multicolumn{3}{c}{\textbf{Fanfics}} &
\multicolumn{3}{c}{\textbf{Random}} \\
\cmidrule(lr){2-4}\cmidrule(lr){5-7}\cmidrule(lr){8-10}
 & Max, tok. &  Gain, tok. &  Information Gain
 & Max, tok. &  Gain, tok. &  Information Gain
 & Max, tok. &  Gain, tok. &  Information Gain \\
\midrule
\textbf{Pythia-160M}  &  80  &  70.9\tiny$\pm$11.0 &  396.4\tiny$\pm$46.0  &  80  &  70.9\tiny$\pm$10.5 &  378.1\tiny$\pm$45.9  &  65  &  61.3\tiny$\pm$6.6  &  500.8\tiny$\pm$38.9 \\
\textbf{Pythia-410M}  &  96  &  81.3\tiny$\pm$12.0 &  431.4\tiny$\pm$51.6  &  96  &  81.2\tiny$\pm$11.6 &  429.8\tiny$\pm$46.2  &  72  &  76.9\tiny$\pm$8.7  &  630.4\tiny$\pm$65.2 \\
\textbf{Pythia-1.4B}  & 160  &  158.0\tiny$\pm$29.1 &  792.8\tiny$\pm$143.4 & 192  &  152.9\tiny$\pm$28.0 &  776.9\tiny$\pm$132.5 & 139  &  144.4\tiny$\pm$17.5 &  1108.2\tiny$\pm$136.2 \\
\textbf{Pythia-2.8B}  & 128  &  150.1\tiny$\pm$50.7 &  740.3\tiny$\pm$234.9 & -  &  - &  - & 141  &  134.5\tiny$\pm$24.7 &  1026.3\tiny$\pm$211.4 \\
\textbf{OPT-1.3B}  & 128  &  132.2\tiny$\pm$23.8 &  712.8\tiny$\pm$143.3 & -  &  - &  - & 116  &  129.3\tiny$\pm$16.4 &  1068.0\tiny$\pm$181.3 \\
\textbf{OLMo-1B}  & 384  &  406.3\tiny$\pm$61.7 &  1901.0\tiny$\pm$254.5 & -  &  - &  - & 249  &  257.3\tiny$\pm$55.0 &  1852.2\tiny$\pm$395.0 \\
\textbf{Sh.LLaMa-1.3B}  & 512  &  383.6\tiny$\pm$38.4 &  1835.1\tiny$\pm$162.9 & -  &  - &  - & 382  &  315.1\tiny$\pm$35.1 &  1893.0\tiny$\pm$210.0 \\
\textbf{Llama-3.2-1B} & 512  &  426.2\tiny$\pm$79.2 &  2119.9\tiny$\pm$364.8 & 512  &  449.6\tiny$\pm$83.7 &  2213.8\tiny$\pm$365.8 & 316  &  294.9\tiny$\pm$64.8 &  2265.2\tiny$\pm$498.7 \\
\textbf{Llama-3.2-3B} & 1024 &  720.3\tiny$\pm$80.2 &  3292.2\tiny$\pm$320.0 & 1024 &  734.1\tiny$\pm$85.0 &  3354.5\tiny$\pm$344.9 & 460  &  456.9\tiny$\pm$72.1 &  3382.6\tiny$\pm$585.2 \\
\textbf{Llama-3.1-8B} & 1568 &  1094.1\tiny$\pm$127.6 &  4865.7\tiny$\pm$546.6 & 1568 &  1071.8\tiny$\pm$168.6 &  4768.9\tiny$\pm$622.6 & 792  &  623.2\tiny$\pm$97.3  &  4541.2\tiny$\pm$758.6 \\
\textbf{Mamba-130M} & 32  &  65.5\tiny$\pm$18.1 &  371.4\tiny$\pm$94.6 & -  &  - &  - & 84  &  62.6\tiny$\pm$17.1 &  535.7\tiny$\pm$121.8 \\
\textbf{Mamba-370M} & 128 &  113.0\tiny$\pm$27.6 &  585.3\tiny$\pm$131.1 & - &  - &  - & 106  &  94.5\tiny$\pm$25.5 &  743.2\tiny$\pm$186.9 \\
\textbf{Mamba-790M} & 256 &  204.2\tiny$\pm$33.0 &  1011.6\tiny$\pm$138.8 & - &  - &  - & 198  &  187.0\tiny$\pm$26.3  &  1421.3\tiny$\pm$179.1 \\
\textbf{Mamba-1.4B} & 512 &  348.9\tiny$\pm$42.9 &  1599.5\tiny$\pm$164.5 & - &  - &  - & 288  &  282.7\tiny$\pm$36.5  &  2062.3\tiny$\pm$257.3 \\
\bottomrule
\end{tabular}
}%
\caption{\textbf{Compression capacity across different text sources and for all evaluated models.}
  We report \emph{Decoding Capacity (in Tokens)} ("Max, tokens" in the Table), \emph{Token Gain}, and \emph{Information Gain} for texts from \emph{PG-19}, \emph{fanfics}, \emph{random}.}
\label{tab:compression_pg_fanfics_random_all_models}
\end{table*}

\section{Understanding the Structure of Compressed Vectors}
\label{app:compressed_vectors_analysis}
To better understand the structure of the space formed by the learned
embeddings, we collect a dataset of embeddings for 64-token sequences
from the GovReport dataset~\cite{huang-etal-2021-efficient}. The
optimization is performed until a reconstruction accuracy of 1.0 is
achieved. Additionally, for each sequence, we compute multiple embeddings
using different random initializations of the \texttt{[mem]} vectors.

\begin{figure}[ht]
    \centering
    \includegraphics[width=0.9\columnwidth]{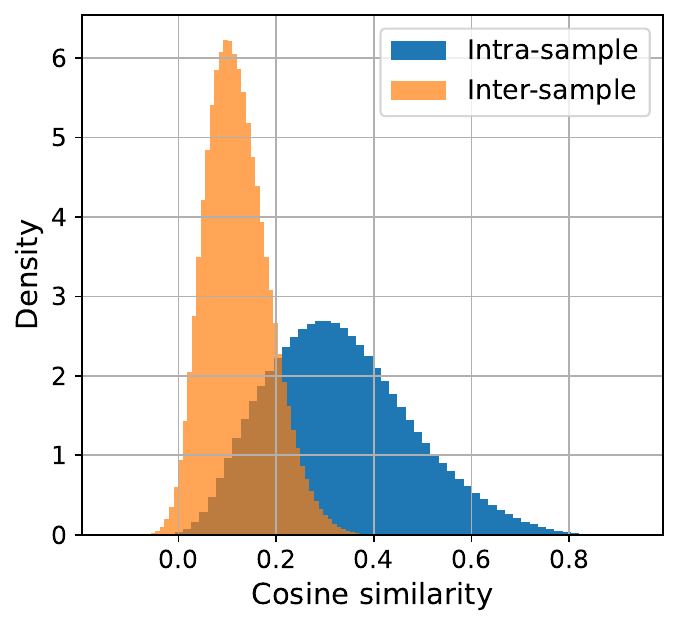}
    \caption{\textbf{Intra/inter-sample embeddings cosine similarity.} Empirical
        probability densities of cosine similarity between intra-sample and inter-sample
        embeddings. Intra-sample similarities are measured between
         of the same sequence of tokens, while inter-sample between different
        ones. Measured on GovReport \cite{huang-etal-2021-efficient} and Sheared-Llama-1.3B \cite{xia2024sheared}.
    }
    \label{fig:cosine_sim}
\end{figure}

First, we observe that the optimization process can yield different
solutions; the resulting vectors for the same text may lie in completely
different parts of the space. To visualize this phenomenon, we plot
histograms of cosine similarity between embeddings of the same text (intra-sample) and between embeddings of different texts
(inter-sample) in~\cref{fig:cosine_sim}. Notably, almost no high cosine
similarities (above 0.8) are observed in the intra-sample case. Moreover,
the intra-sample similarities significantly overlap with the inter-sample
ones, implying that the embeddings are considerably scattered throughout
the space.

\begin{figure}[ht]
    \centering
    \includegraphics[width=0.9\columnwidth]{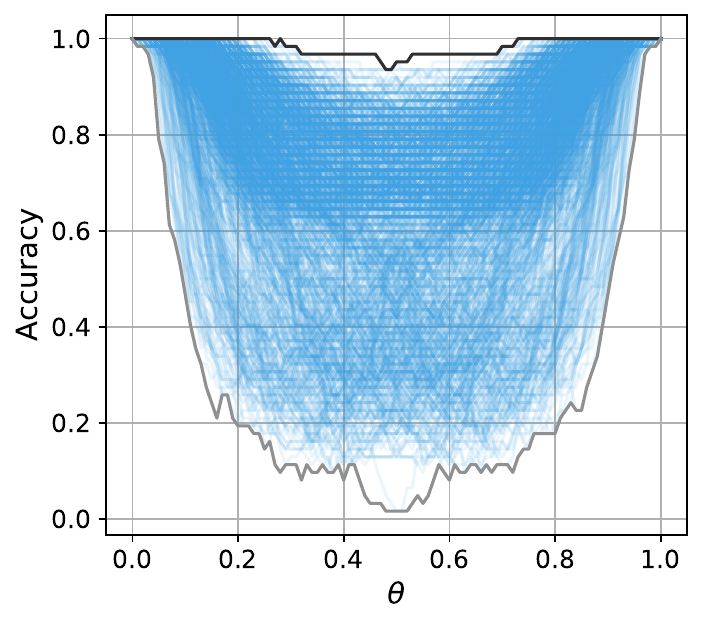}
    \caption{\textbf{Intra-sample Interpolation Accuracies.} Interpolation lines 
    are provided for all pairs between 32 embeddings of the same input sequence. 
    All interpolation lines are printed with high transparency 
    to show denser regions. Grey lines depict minimums and maximums of the 
    accuracy for a given interpolation parameter $\theta$.
    }
    \label{fig:interp}
\end{figure}

Although the embeddings appear to be spread out, one might hope they form
a basin in which all linear interpolations between vectors would yield
perfect reconstruction. To test this, we computed the reconstruction accuracy
along linear interpolation trajectories between embeddings of the same
sequence. However, in all cases we examined, errors were present along the
interpolation trajectory (see~\cref{fig:interp}). Thus, the embeddings
obtained by the proposed procedure do not form a continuous basin.

These observations have several implications:
\begin{enumerate}
    \item A lossless compression algorithm ideally assigns a unique decoding
    to each vector; multiple valid embeddings for the same object limit the
    achievable compression rate.
    \item The spread and entangled structure of the embeddings may render them
    less useful as representations.
    \item This non-unique, scattered structure could make it more challenging
    to extract important information when these compressed representations are
    used as context in an LM.
\end{enumerate}

\section{Data Compression Analysis}
\label{app:data_compression}
In the present work, we focus on exploring phenomena that could be potentially helpful in building more efficient LLMs. Despite our method can serve as a general-purpose compression approach, we currently treat lossless compression as a tool to better understand LLMs hidden state capacity in contrast to achieving the highest possible compression rate.

We ran experiments on three datasets (PG-19, fanfics, and randomly sampled sequences of words) using zlib, bz2, lzma, pure Huffman coding, and Arithmetic Coding(AC) with LM. We report their mean compression ratios (original text size in bits / compressed size) in \cref{tab:classic_compression_rate_comparison}.

\begin{table}[h]
\begin{center}
\resizebox{\linewidth}{!}{%
\begin{tabular}{llll}
\hline
\multicolumn{1}{c}{\textbf{Method}} & \multicolumn{1}{c}{\textbf{PG19}} & \multicolumn{1}{c}{\textbf{Fanfics}} & \multicolumn{1}{c}{\textbf{Random}} \\
\hline
zlib                                & 2.28 ± 0.16                       & 2.34 ± 0.07                          & 1.80 ± 0.06                         \\
bz2                                 & 2.46 ± 0.16                       & 2.56 ± 0.09                          & 1.94 ± 0.11                         \\
lzma                                & 2.28 ± 0.16                       & 2.33 ± 0.07                          & 1.86 ± 0.13                         \\
Huffman                             & 1.81 ± 0.07                       & 1.86 ± 0.04                          & 1.77 ± 0.01                         \\
\hline
AC, pythia-160m      & 6.77 ± 0.75                       & 6.73 ± 0.41                          & 2.83 ± 0.08                         \\
\hline
\end{tabular}
}
\end{center}
\caption{Compression ratios (in bits) comparison for classic compression algorithms and arithmetic coding (AC) using pythia-160m for a range of corpora.}
\label{tab:classic_compression_rate_comparison}
\end{table}

We see that conventional entropy coders can compress texts by about a factor of two, and combining Arithmetic Coding with pythia-160m can lead to much higher ratios: 6-7x. With our approach we can encode up to 1568 tokens into a single 4096-dimensional bfloat16 vector (using LLaMA-3.1-8B) and reconstruct them losslessly. If we treat a single 4096-dimensional vector as a "compressed file" and compute (original text size as string) / (size of vector), the ratio actually appears to be about 0.8x, so it does not take less bits on disk than a raw text.

Importantly, unlike conventional compression algorithms that output arbitrary bitstreams, our approach must encode the text into a single vector, which LLMs interpret as an embedding. This constraint is stricter than simply minimizing file size, since the output must lie within the input space of the model. Our focus, therefore, is on analyzing representational capacity within the input space of large language models. For instance, a single 4096-dimensional vector in LLaMA-3.1-8B can guide the model to generate up to 1568 tokens exactly. Reconstructing 1568 tokens from just one vector results in a 1568x reduction in the number of embeddings that would otherwise be used to represent the text.

Our goal is not to outperform standard compressors in bits-per-byte efficiency, but rather to show that LLMs can store significant amounts of text with only a single embedding.

\end{document}